\documentclass[conference]{IEEEtran}
\IEEEoverridecommandlockouts
\usepackage{cite}
\usepackage[utf8]{inputenc}
\usepackage[english,vietnamese]{babel}
\usepackage{amsmath,amssymb,amsfonts}
\usepackage{algorithmic}
\usepackage{graphicx}
\usepackage{textcomp}
\usepackage{xcolor}
\usepackage{url}
\usepackage{hyperref}

\DeclareMathOperator*{\argmax}{argmax}
\def\BibTeX{{\rm B\kern-.05em{\sc i\kern-.025em b}\kern-.08em
    T\kern-.1667em\lower.7ex\hbox{E}\kern-.125emX}}

\begin{document}

\title{Neural sequence labeling for Vietnamese \\POS Tagging and NER}

\author{
    \IEEEauthorblockN{Anh-Duong Nguyen}
    \IEEEauthorblockA{\textit{Hanoi Univ of Science and Technology} \\
    Hanoi, Vietnam \\
    20160840@student.hust.edu.vn}
    \and
    \IEEEauthorblockN{Kiem-Hieu Nguyen}
    \IEEEauthorblockA{\textit{Hanoi Univ of Science and Technology} \\
    Hanoi, Vietnam \\
    hieunk@soict.hust.edu.vn}
    \and
    \IEEEauthorblockN{Van-Vi Ngo}
    \IEEEauthorblockA{\textit{VCCorp} \\
    Hanoi, Vietnam \\
    vingovan@admicro.vn}
}
\maketitle

\selectlanguage{english}

\begin{abstract}
    This paper presents a neural architecture for Vietnamese sequence labeling tasks including part-of-speech (POS) tagging and named entity recognition (NER). We applied the model described in \cite{lample-EtAl:2016:N16-1} that is a combination of bidirectional Long-Short Term Memory and Conditional Random Fields, which rely on two sources of information about words: character-based word representations learned from the supervised corpus and pre-trained word embeddings learned from other unannotated corpora. Experiments on benchmark datasets show that this work achieves state-of-the-art performances on both tasks - 93.52\% accuracy for POS tagging and 94.88\% F1 for NER. Our sourcecode is available at  \href{https://github.com/duongna21/VNsequencelabeling}{here}.
\end{abstract}

\begin{IEEEkeywords}
sequence labeling
, part-of-speech tagging, named-entity recognition, character-level knowledge, syntactic information
\end{IEEEkeywords}

\section{Introduction}
Linguistic sequence tagging is a well studied yet challenging problem in natural language processing. Most traditional high performance sequence tagging systems apply supervised learning techniques such as Hidden Markov Models, Maximum entropy Markov models, and Conditional Random Fields (CRFs) \cite{ratinov2009design}\cite{passos2014lexicon}\cite{luo2015joint}, which require large amounts of hand-crafted features and domain-specific knowledge. For example, sequence taggers benefit from carefully constructed word-shape features; external language-specific knowledge resources such as gazetteers are also widely used for solving NER task. However, such task-specific knowledge are inherently limited and can be costly to develop, making these models difficult to adapt to new languages or new domains.

Recently, deep learning has been extensively applied to sequence tagging in many languages, and the focus has shifted from feature engineering to designing and implementing effective neural network architectures \cite{lample-EtAl:2016:N16-1} \cite{ma-hovy:2016:P16-1}\cite{huang2015bidirectional}.

There are several work applying deep learning to sequence tagging in Vietnamese \cite{Pham:2017b}\cite{nguyen2014rdrpostagger}\cite{Pham:2017a}. However, they have not achieved accuracy that far beyond that of classical machine learning methods. In this paper, we demonstrate the success of applying a neural architecture for sequence labeling to Vietnamese. The model requires no task-specific resources, hand-engineered, or data pre-processing beyond pre-trained word embeddings on unannotated corpora \cite{lample-EtAl:2016:N16-1}. The model first uses bidirectional Long-Short Term Memory units (Bidirectional LSTM) to learn character embeddings. Next, character representation is concatenated with the pre-trained word embeddings and then applying dropout \cite{srivastava2014dropout} to encourage the model to learn to trust both sources of evidence, then feed them into a bidirectional LSTM to capture context information of each word. On top of bidirectional LSTM, sequential CRFs are used to jointly decode labels for the whole sentence.  Experiments on benchmark datasets show that we obtain state-of-the-art results on both tasks - 93.52\% accuracy for POS tagging and 94.88\% F1 for NER.

The main contributions of this paper are three-fold:
\begin{itemize}
    \item We demonstrate the effectiveness of leveraging character-level knowledge for language representations.
    \item We apply a neural sequence labeling system for Vietnamese that outperforms traditionally machine learning methods and other deep learning models.
    \item We release all code to the research community.
    
\end{itemize}
\section{Related Work}
As mentioned earlier, we can basically categorize main approaches to sequence tagging into feature-based machine learning models and deep learning models.

The first approach includes traditional models that rely heavily on hand-crafted features and domain-specific knowledge. For Vietnamese, previously published sequence labeling systems used traditional machine learning methods such as CRFs \cite{tran2013vtools} and MEMM \cite{le2010empirical}, \cite{tran2013vtools}. Recently, several participants at the VLSP 2016 workshop for NER task use MEMM \cite{le2016vietnamese} and CRFs \cite{le2016named} to solve this problem.

In deep learning approach, our work is closely related to NNVLP \cite{Pham:2017b}. Being motivated by CNN-LSTM-CRF architecture that achieves state-of-the-art performance in many languages \cite{ma-hovy:2016:P16-1}, NNVLP applied this architecture to Vietnamese. In our work, instead of CNNs, we use LSTM to represent character embeddings as in \cite{lample-EtAl:2016:N16-1}.

\section{Backgrounds on LSTM and CRFs}
\subsection{Recurrent Neural Networks}
Recurrent Neural Networks (RNNs) are a family of artificial neural network for processing sequential data \cite{rumelhart1985learning}. They take a sequence of vectors $\boldsymbol{x}$ as input and return another sequence $\boldsymbol{h}$ representing some information about the sequence at every step in the input. An RNN introduces the connection the previous hidden state and current hidden state. At time step $t$, the value in the hidden layers and output layers $\boldsymbol{o}$ are computed as follows:
\begin{align}
    &\boldsymbol{h}_t = f(\boldsymbol{Ux}_t+\boldsymbol{Wh}_{(t-1)})\\
    &\boldsymbol{o}_t=\boldsymbol{g}(\boldsymbol{Vs}_t), 
\end{align}

where $\boldsymbol{U}, \boldsymbol{W}$ and $\boldsymbol{V}$ are the parameters computed in training time. $f$ usually is a nonlinear function such as $\text{tanh}$ or $\text{ReLU}$, and $g$ is a softmax function.

RNNs maintain a memory based on history information which enables the model can, in theory, predict the current output conditioned on long distance features. In practice they fail to do, and tend to be biased towards their most recent inputs, due to the gradient vanishing/exploding problems \cite{bengio1994learning} \cite{pascanu2013difficulty}. 

\subsection{LSTMs} 
LSTMs \cite{hochreiter1997long} are variants of RNNs which have been designed to deal with these gradient vanishing problems by replacing the RNN’s hidden layers by purpose-built memory cells, and have been shown to effectively capture long-range dependencies. An LSTM unit is composed of several gates which control the proportions of information to forget and to pass on to the next time. Thus they can be better at finding and exploiting long range dependencies in the data. Formally, the memory cell is updated at time $t$ as follow:
\begin{align}
    &\boldsymbol{f}_t=\sigma(\boldsymbol{W}_f\boldsymbol{h}_{t-1}+\boldsymbol{U}_f\boldsymbol{x}_t+\boldsymbol{b}_f)\\
    &\boldsymbol{i}_t=\sigma(\boldsymbol{W}_i\boldsymbol{h}_{t-1}+\boldsymbol{U}_i\boldsymbol{x}_t+\boldsymbol{b}_i)\\
    &\boldsymbol{\tilde{c}_t}=\text{tanh}(\boldsymbol{W}_c\boldsymbol{h}_{t-1}+\boldsymbol{U}_c\boldsymbol{x}_t+\boldsymbol{b}_c)\\
    &\boldsymbol{c}_t=\boldsymbol{f}_t\odot\boldsymbol{c}_{t-1}+\boldsymbol{i}_t\odot\boldsymbol{\tilde{c}_t}\\
    &\boldsymbol{o}_t=\sigma(\boldsymbol{W}_o\boldsymbol{h}_{t-1}+\boldsymbol{U}_o\boldsymbol{x}_t+\boldsymbol{b}_o)\\
    &\boldsymbol{h}_t=\boldsymbol{o}_t\odot\text{tanh}(\boldsymbol{c}_t),
\end{align}
where $\sigma$ is the logistic sigmoid function and $\odot$ is the element-wise product. $\boldsymbol{x}_t$ is the input vector at time $t$, and $\boldsymbol{h}_t$ is the hidden state vector capturing the useful information at current time (and all the previous time steps). $\boldsymbol{f}$, $\boldsymbol{i}$, $\boldsymbol{o}$ and $\boldsymbol{c}$ are the forget gate, input gate, output gate and cell vectors respectively, with $\boldsymbol{U}_f$, $\boldsymbol{U}_i$, $\boldsymbol{U}_o$, $\boldsymbol{U}_c$ are weight matrices of these gates for input $\boldsymbol{x}_t$ and $\boldsymbol{W}_f$, $\boldsymbol{W}_i$, $\boldsymbol{W}_o$, $\boldsymbol{W}_c$ are the weight matrices for hidden state $\boldsymbol{h}_t$. 

\subsection{Bidirectional LSTM} 
\begin{figure}
\centering
\includegraphics[width=0.5\textwidth]{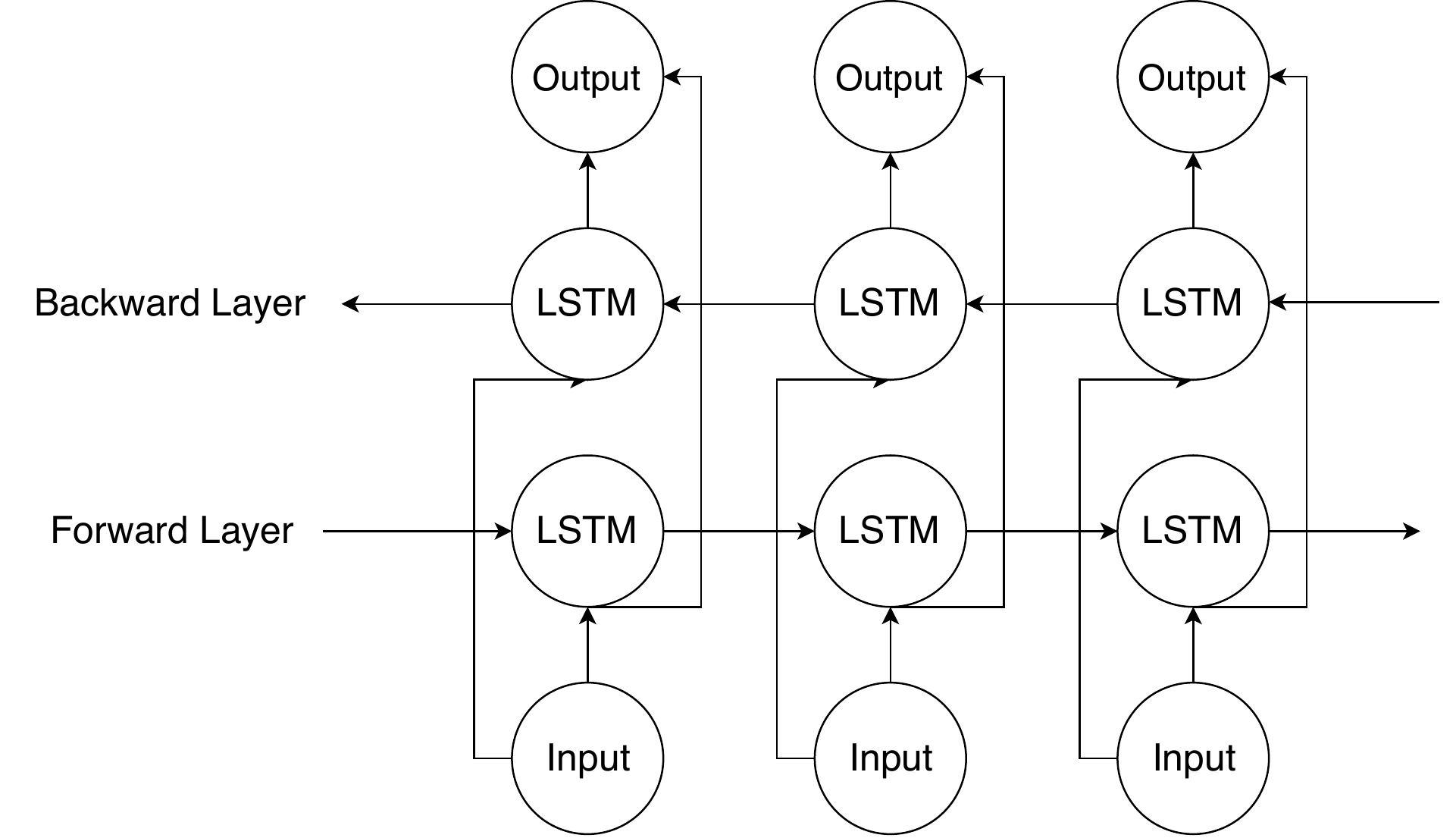}
\caption{Bidirectional LSTM}
\label{fig:LSTM}
\end{figure}

An LSTM takes information in a sequence only from past. However, in many sequence tagging tasks, retrieving information from both past (left) and future (right) contexts is beneficial. A straightforward way to do this is using a bidirectional LSTM network (figure~\ref{fig:LSTM}) as proposed in \cite{graves2005framewise}. generates a representation of the right context that can be achieved by using a second LSTM reading the same sequence in reverse. Then the final output is obtained by concatenating its left and right context representations.

\subsection{Bidirectional LSTM-CRFs}
For many sequence labeling tasks, a very simple but surprisingly effective method is to consider the correlations between labels in neighborhoods and jointly decode the best chain of labels for a given input sequence. For example, in NER task with BIO2 annotation \cite{sang1999representing}, B-LOC is certainly followed by I-LOC, and I-PER cannot follow B-ORG. Therefore, instead of decoding each label independently, we decode them jointly using a linear-chain first order CRF \cite{lafferty2001conditional}. CRFs is a family of discriminative probabilistic framework, which directly model conditional probabilities of a tag sequence given a word sequence. Formally, the score of the input sentence $\boldsymbol X = (\boldsymbol x_1, \boldsymbol x_2,\dots, \boldsymbol x_n)$, where $\boldsymbol x_i$ is the vector of the $i^\textit{th}$ word, associated with the sequence of tag $\boldsymbol{y} = (y_1, y_2,\dots, y_n)$ is defined to be
$$s(\boldsymbol{X, y})=\sum_{i=0}^{n}A_{y_{i},y_{i+1}}+\sum_{i=1}^{n}P_{y_i, i},$$
where $A_{i,j}$ denotes the score of a transition from the tag $i$ to the tag $j$, $y_0$ and $y_{n+1}$ are the $\textit{start}$ and $\textit{end}$ tags of a sentence. $P_{y, i}$ represents the score of the tag $y$ of the $i_{th}$ word in a sentence outputted by the bidirectional LSTM network. 

Then, the probability of the tag sequence is given with the following form
$$p(\boldsymbol{y}\vert\boldsymbol{X})=\frac{\exp{(s(\boldsymbol{X}, \boldsymbol{y}}))}{\sum_{\boldsymbol{y}'\in\boldsymbol{Y_{X}}}\exp({s(\boldsymbol{X}, \boldsymbol{y}'))}}$$
where $\boldsymbol{Y}_{\boldsymbol{X}}$ is the set of all possible tag sequences for a sentence $\boldsymbol{X}$. 
For CRF training, the parameters can be estimated by maximizing the log-probability of the correct tag sequence $\log({p(\boldsymbol{y\vert\boldsymbol{X}}))}$, and the output sequence $\boldsymbol{y}^*$ is obtained by maximizing the score 
$$\boldsymbol{y}^{*}=\argmax_{\boldsymbol{y}'\in\boldsymbol{Y_{\boldsymbol{X}}}}s(\boldsymbol{X},\boldsymbol{y}')$$
The decoding stage in CRFs (i.e. linear first-order CRFs in our work) can be efficiently solved by Viterbi algorithm.

\section{Bidirectional LSTM-CRFs with character embeddings for sequence labeling}
\subsection{Bidirectional LSTM-CRFs with character embeddings for POS Tagging}
\begin{figure}
\centering
\includegraphics[width=0.5\textwidth]{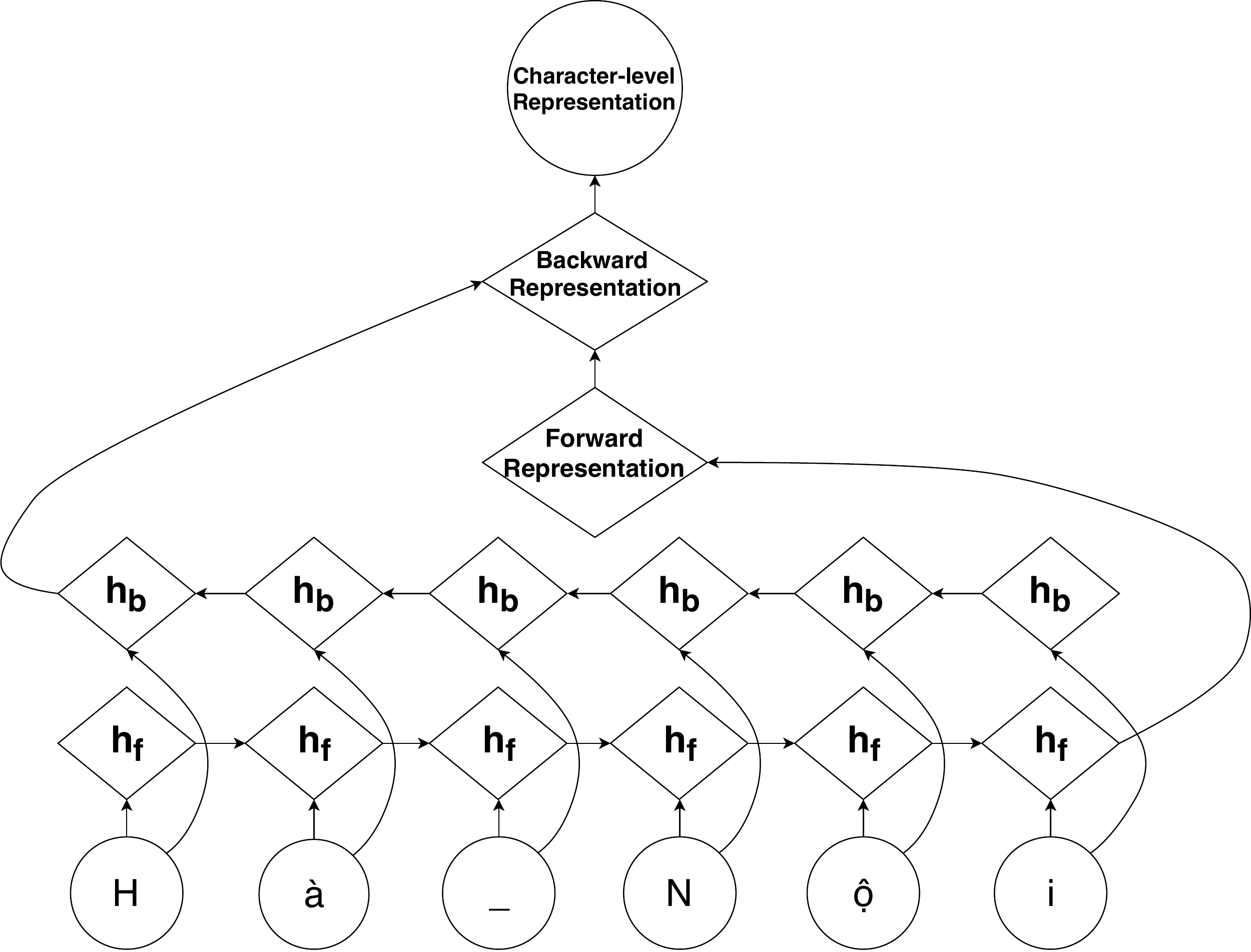}
\caption{The character embeddings of the word \selectlanguage{vietnamese} ``Hà\_Nội'' \selectlanguage{english} are given to a bidirectional LSTMs for extracting character-level word features. Character-level representation of one word is the concatenation of its forward and backward representations}
\label{fig:char}
\end{figure}

\textbf{Character-based model of word}.
Recently, character-level knowledge has been leveraged and empirically verified to be helpful in sequence labeling tasks. \cite{lample-EtAl:2016:N16-1} has shown that LSTM is an effective approach to extract morphological information (instead of hand-engineering prefix and suffix information about words) from characters of words. Firstly, we randomly initialize an embedding for every character. The character embeddings corresponding to characters in a word are given to a forward and a backward LSTM. For each word, the character-level word embedding is the concatenation of two last hidden states from forward and backward layers of the Bidirectional LSTM as described in Figure~\ref{fig:char}. Then, this character-based word representation is concatenated with the pre-trained word embedding to obtain the word representation.

\begin{figure}
\centering
\includegraphics[width=0.5\textwidth]{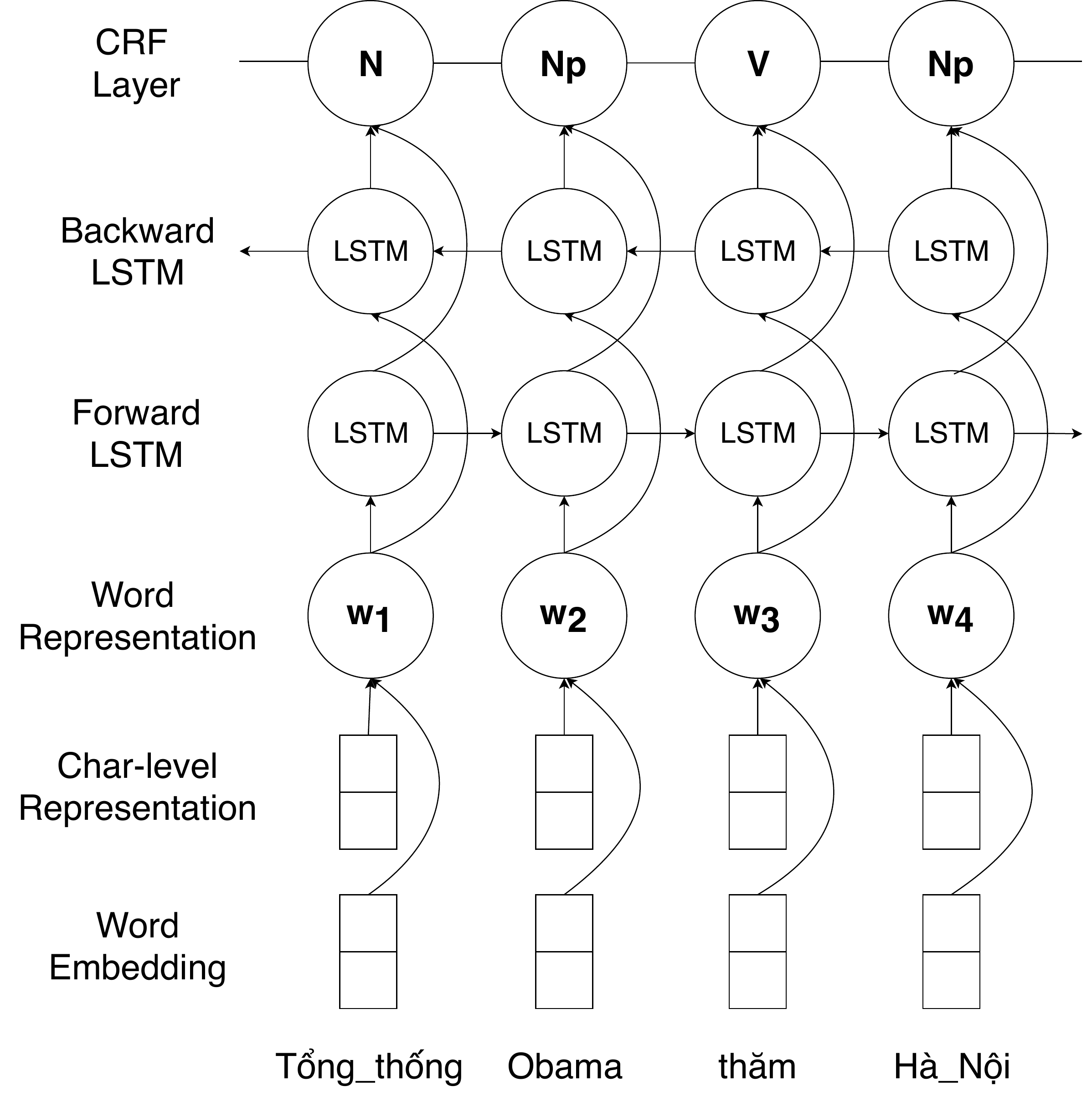}
\caption{Bidirectional LSTM-CRFs architecture for POS tagging. }
\label{fig:pos_tagging}
\end{figure}

\textbf{Pre-trained word embeddings}.
In this work, we use available word embedding set provided by \cite{Pham:2017b}, that trained from 7.3GB of 2 million articles collected through a Vietnamese news portal by word2vec toolkit. For words that do not have an embedding, a vector called unknown (UNK) embedding is used instead. The UNK embedding is created by random vectors sampled uniformly from the range $[-\sqrt{\frac{3}{dim}},+\sqrt{\frac{3}{dim}}]$ where \textit{dim} is the dimension of word embeddings \cite{he2015delving}. Here the number of dimensions for word embedding is 300. Dropout layers are applied on both the input and output vectors of Bidirectional LSTM to prevent the models from depending on one representation or the other too strongly. Experimental results have shown the significant effectiveness of the use of dropout in training. Finally, the output vectors of Bidirectional LSTM are fed to a CRF layer to jointly decode the best label sequence. Figure~\ref{fig:pos_tagging} presents the architecture of this approach.
\subsection{Bidirectional LSTM-CRFs with character embeddings for NER.}
\begin{figure}
\centering
\includegraphics[width=0.5\textwidth]{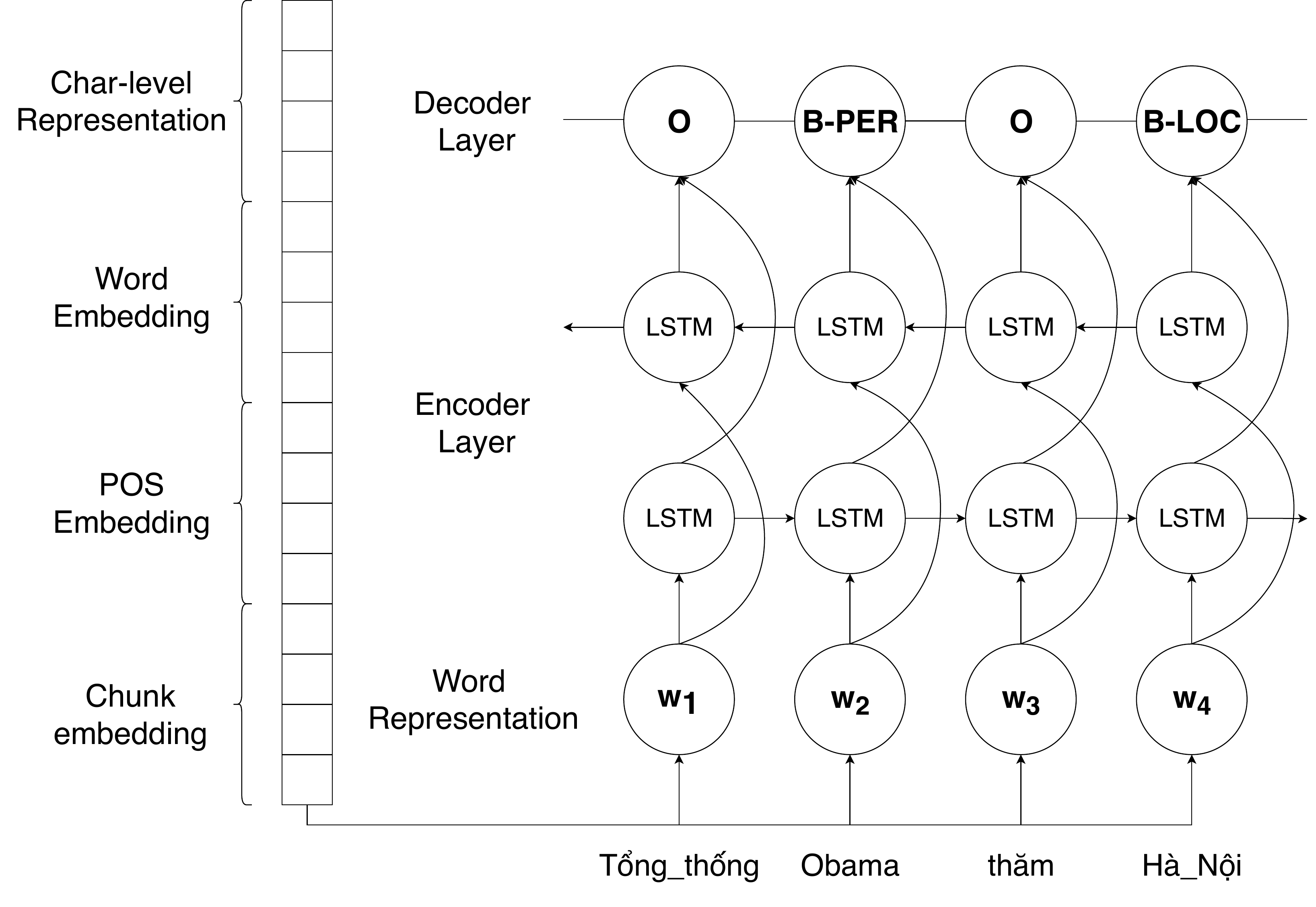}
\caption{Bidirectional LSTM-CRFs architecture for NER}
\label{fig:ner}
\end{figure}

For NER task, we apply the same system as in the POS tagging task. Furthermore, we utilizes other information such as POS tags or chunk tags that are available in the dataset to provide additional syntactic information. To do this, we encode every POS (and chunk) tag as a one-hot vector whose length is equal to the number of POS tags (and chunk tags). For example: 
\begin{itemize}
    \item N:  $1000...000$
    \item V:  $0100...000$
    \item R:  $0000...010$
\end{itemize}
Then, the concatenation of these one-hot vectors and word representation are fed into Bidirectional LSTM as described in Figure~\ref{fig:ner}.

\section{Experiments}
\subsection{Datasets}
We use the following benchmark datasets in our experiments: VietTreebank (VTB) \cite{nguyen2009building} for POS tagging, and VLSP shared task 2016 corpus for NER task. While the NER dataset has been originally released for evaluation, the POS tagging data set are not previously divided. Therefore, we use cross-validation to evaluate POS tagging. Table~\ref{tab:datasets} provides some statistics about these corpus. The dataset provided by VLSP 2016 organizers consist of four columns of word, POS, chunk and NER tag respectively.

\begin{table}[]
    \centering
    \caption{Dataset statistics}
    \begin{tabular}{llrr}
        \hline
         Dataset & Task & \#sent & \#word \\
         \hline
         VTB & POS tagging & 10,383 & 221,464 \\
         VLSP 2016 - train & NER & 14,861 & 325,686 \\
         VLSP 2016 - dev & NER & 2,000 & 43,706 \\
         VLSP 2016 - test & NER & 2,831 & 66,097 \\
        \hline
    \end{tabular}
    \label{tab:datasets}
\end{table}

\subsection{Hyper-parameters}
Table~\ref{tab:hyperparam} summarizes the chosen hyper-parameters for all experiments.
\begin{table}[]
    \centering
    \caption{The model Hyper-parameters}
    \begin{tabular}{lr}
        \hline
         Hyper-parameter & Value \\
         \hline
         character dimension & 100 \\
         word dimension & 300 \\
         hidden size char & 100 \\
         hidden size word & 150 \\
         update function & Adam \\
         learning rate & 0.0035 \\
         learning decay rate & 0.005 \\
         dropout rate & 0.35 \\
         batch size & 8 \\
        \hline
    \end{tabular}
    \label{tab:hyperparam}
\end{table}
Hyper-parameters were tuned on the development set of NER. On the hand, we used hyper-parameters tuned for NER to learn POS tagging.

We set the character embedding dimension at 100, the word embedding dimension at 300, the dimension of hidden states of the char-level LSTMs at 100 and the dimension of hidden states of the word-level LSTMs at 150. We optimize parameters using Adaptive Moment Estimation (Adam) optimizer \cite{kingma2014adam} with mini-batch size 8. We choose an initial learning rate of $\eta_0=0.003$, and the learning rate is updated on each epoch of training as $\eta_t = \eta_0/(1+\alpha t)$, with decay rate $\alpha = 0.05$ and $t$ is the number of epoch completed \cite{ma-hovy:2016:P16-1}. Inputs for POS tagging task are gold word segmentation, and for NER task are gold word segmentation, POS tags, and chunks. As mention before, dropout method is applied on both the input and output vectors of the second Bidirectional LSTM to prevent overfitting. We fix the dropout rate to $0.35$. We use early stopping based on performance on development sets in which the convergence is reached after around $25$ epochs.

\subsection{Experimental Results}

For POS tagging, we compare our work with the following supervised taggers:
\begin{itemize}
    \item NNVLP \cite{Pham:2017b} is a deep learning method based on Bidirectional LSTM-CRFs architecture with character embeddings from CNNs.
    \item RDRPOSTagger  \cite{nguyen2014rdrpostagger} learns tagging rules from annotation using Ripple-Down Rules framework.
    \item VNTagger \cite{le2010empirical} learns a Maximum Entropy model using handcraft features.
\end{itemize}

For NER, we compare our work with the following supervised methods:
\begin{itemize}
    \item Feature-rich CRFs \cite{minh2018feature}: As the name mentions, this system learn CRFs model with a rich set of features including surrounding n-grams and word shape features.
    \item NNVLP \cite{Pham:2017b}: This model is identical to the model for POS tagging except the inclusion of POS tags and chunk embeddings.
\end{itemize}

We conduct experiments and compare the performance of our system and several published systems on POS tagging and NER task.  In experiments, we use micro-averaged F1 score, the official evaluation metric in CoNLL 2003 \cite{tjong2003introduction} as the evaluation measure for NER task. The results on these task are plotted in Table~\ref{tab:POS_tagging} and Table~\ref{tab:NER} respectively. These tables compares the Vietnamese POS tagging and NER results of our system with results reported in prior work, using the same experimental setup. In particular, our system achieves an accuracy of 93.52\% for POS tagging task and a F1 score of 94.88\% for NER task, which significantly outperform previous state-of-the-art work on Vietnamese. Experimental results showed significant improvement on both tasks, especially for NER, gained by leveraging character-level knowledge of words. We also observed that NER task benefits from syntactic features including POS and chunk information, which aligns with the results in \cite{pham2017importance}.  

\begin{table}[]
    \centering
    \caption{POS tagging performance on VTB}
    \begin{tabular}{lrr}
        \hline
         \textbf{Method} & \textbf{Accuracy} & \textbf{Evaluation Method} \\
         \hline
         NNVLP \cite{Pham:2017b} & 91.92 \\
         RDRPOStagger \cite{nguyen2014rdrpostagger} & 92.59 & 5-fold cross-validation \\ 
         BiLSTM-CRFs w.o char  & 91.74 \\
         BiLSTM-CRFs  & \textbf{92.98} \\
         \hline
         VNTagger \cite{le2010empirical} & 93.40 \\
         BiLSTM-CRFs w.o char  & 92.22 & 10-fold cross-validation\\
         BiLSTM-CRFs  & \textbf{93.52} \\
        \hline
    \end{tabular}
    \label{tab:POS_tagging}
\end{table}

\begin{table}[]
    \centering
    \caption{NER performance on VLSP 2016 dataset}
    \begin{tabular}{lrrrr}
        \hline
         \textbf{Method} & \textbf{P} & \textbf{R} & \textbf{F1} & \textbf{F1} \\
         & & & & \textbf{(w.o char)} \\
         \hline
         Feature-rich CRFs \cite{minh2018feature} & 93.87 &  93.99 & 93.93 & - \\
         NNVLP \cite{Pham:2017b} & 92.76 & 93.07 & 92.91 & - \\
         \hline 
         BiLSTM-CRFs  & 90.97 & 87.52 & 89.21 & 76.43 \\
         BiLSTM-CRFs + POS & 90.90 & 90.39 & 90.64  & 86.06 \\
         BiLSTM-CRFs + Chunk & 95.24 & 92.16 & 93.67 & 87.13 \\
         BiLSTM-CRFs + POS + Chunk & \textbf{95.44} & \textbf{94.33} & \textbf{94.88} & 91.36\\
        \hline
    \end{tabular}
    \label{tab:NER}
\end{table}

As our model significantly outperforms NNVLP which applies the same Bidirectional LSTM-CRFs framework and the same pre-trained word embedding set, we take a closer look at their model. Firstly, their neural networks have different hyper-parameter values. Secondly, their CNNs use only one window size of three subsequent characters. Therefore, it could possibly be that our hyper-parameters were better tuned on development set; or their CNNs-based character embeddings should be further improved, for instance by combining several window sizes; or LSTMs, which capture long-range dependencies and cross-syllable dependencies, are better for character embeddings of Vietnamese words. In the future, we are going to investigate more on this.

\section{Conclusions}
In this work, we have applied a deep neural network model for Vietnamese sequence tagging, which obtains state-of-the-art performance. Experiments on benchmark data sets for Vietnamese sequence labeling tasks showed the effectiveness of training character-based model of word by bidirectional LSTM in language representation serving sequence labeling tasks, that outperform models using CNN despite requiring less cost for building. We have also shown that inserting syntactic features into the representation of a word improves accuracy significantly, especially information about chunk.

\section*{Acknowledgment}
This work is supported by VCCorp. We would like to thank the NNVLP \cite{Pham:2017b} team for publishing the pre-trained word embedding set that we used during training and evaluating stage of our model.

\bibliography{bibliography}
\bibliographystyle{IEEEtran}

\end{document}